\documentclass[]{article}
\usepackage{proceed2e}

\usepackage{times}

\usepackage{hyperref}

\usepackage{url}

\usepackage{bm}
\usepackage{amsmath,amssymb}
\usepackage{xspace}
\usepackage{subfigure, graphicx}
\usepackage[colon,numbers]{natbib}
\usepackage{bm}
\usepackage{subfigure}
\usepackage{dsfont}
\usepackage{psfrag}
\usepackage{rotating}
\usepackage{epstopdf}

\usepackage{algorithm}
\usepackage{algorithmic}

\usepackage{color}

\floatstyle{ruled}
\newfloat{algo}{thp}{lop}
\floatname{algo}{Algorithm}

\newcommand{\softmax}{\mathop{\varsigma}}
\newcommand{\OmegaSpace}{\Omega}

\newcommand{\TauSpace}{\mathcal{T}}
\newcommand{\f}{f}

\newcommand{\q}{q}

\renewcommand{\a}{a}

\renewcommand{\d}{d}
\newcommand{\w}{w}

\newcommand{\x}{x}
\newcommand{\y}{y}

\newcommand{\transp}{^{T}}

\newcommand{\grad}{\nabla}

\newcommand{\U}{\bm{U}}

\renewcommand{\P}{P}
\newcommand{\Q}{Q}

\newcommand{\Expectation}[2]{\mathbb{E}_{#1}\left[#2\right]}

\newcommand{\Var}[2]{\textnormal{Var}_{#1}\left[#2\right]}

\providecommand{\U}[1]{\protect\rule{.1in}{.1in}}
\newcommand{\be}{\begin{equation}}
\newcommand{\ee}{\end{equation}}
\newcommand{\bd}{\begin{definition}}
\newcommand{\ed}{\end{definition}}
\newcommand{\ba}{\begin{algorithm}}
\newcommand{\ea}{\end{algorithm}}
\newcommand{\br}{\begin{problem}}
\newcommand{\er}{\end{problem}}
\newcommand{\bex}{\begin{example}}
\newcommand{\eex}{\end{example}}
\newcommand{\bt}{\begin{theorem}}
\newcommand{\et}{\end{theorem}}

\newtheorem{theorem}{Theorem}

\newtheorem{definition}[theorem]{Definition}
\newtheorem{example}[theorem]{Example}

\newtheorem{problem}[theorem]{Problem}

\def\ifa{\iffalse}
\def\ifappendix{\iftrue}

\newcommand{\expf}{\gamma} % Bottou notation seems not very popular and risk of confusing with Jacobian, so sticking with Guillaume's notation makes sense
\newcommand{\param}{w} % Bottou notation
\newcommand{\paramspace}{\mathcal{W}}  % Bach notation (also for 'H'ypothesis space)
\newcommand{\sample}{x}  % Bottou and Bach notation
\newcommand{\samplespace}{\mathcal{X}}
\title{Online Learning to Sample}

\author{
 Guillaume Bouchard\thanks{\ \ Work made while at Xerox.} \\
 University College London\\
 \texttt{g.bouchard@cs.ucl.ac.uk} \\
 \And
 Th\'eo Trouillon \\
 Xerox Research Center Europe\\
 UGA - LIG\\
 \texttt{theo.trouillon@xrce.xerox.com}\\
 \AND
 Julien Perez \\
 Xerox Research Center Europe \\
 \texttt{julien.perez@xrce.xerox.com} \\
 \And
 Adrien Gaidon \\
 Xerox Research Center Europe \\
 \texttt{adrien.gaidon@xrce.xerox.com} \\
}

\graphicspath{{pics/}}
\begin{document}

\iftrue

\maketitle

\begin{abstract}
Stochastic Gradient Descent (SGD) is one of the most widely used techniques for online optimization in machine learning.
In this work, we accelerate SGD by adaptively
learning how to sample the most useful training examples at each time step. First, we show that SGD can be used to learn the best possible sampling distribution of an importance sampling estimator. Second, we show that the
sampling distribution of an SGD algorithm can be estimated online by incrementally minimizing the variance of the gradient.
The resulting algorithm --- called Adaptive Weighted SGD (AW-SGD) --- maintains a set of parameters to optimize, as well as a set of parameters to sample learning examples.
We show that AW-SGD yields faster convergence in three different applications: (i) image classification with deep features, where the sampling of images depends on their labels, (ii) matrix factorization, where rows and columns are not sampled uniformly, and (iii) reinforcement learning, where the optimized and exploration policies are estimated at the same time, where our approach corresponds to an off-policy gradient algorithm.
\end{abstract}

\vspace*{3mm}
\section{Introduction}

In many real-world problems, one has to face intractable integrals, such as
averaging on combinatorial spaces or non-Gaussian integrals. Stochastic
approximation is a class of methods introduced in 1951 by Herbert Robbins and
Sutton Monro~\cite{RobbinsMonro1951} to solve intractable equations by using a
sequence of approximate and random evaluations. Stochastic Gradient
Descent~\cite{bottou-98x} is a special type of stochastic approximation method
that is widely used in large scale learning tasks thanks to its good
generalization properties~\cite{bottou-bousquet-2011}.

Stochastic Gradient Descent (SGD) can be used to minimize functions of the
form:
\begin{eqnarray}
\expf(\param) &:=& \Expectation{\sample \sim P}{f(\sample;\param)}
                = \int_{\samplespace} f(\sample;\param) d\P(\sample)
\label{eq:sgdObjective}
\end{eqnarray}
where $P$ is a known fixed distribution and $f$ is a function that maps
$\samplespace \times \paramspace$ into $\Re$, i.e. a family of functions on the
metric space $\samplespace$ and parameterized by $\param \in \paramspace$.
SGD is a stochastic approximation method that consists in doing approximate
gradient steps equal on average to the true gradient $\grad_\param
\expf(\param)$~\cite{bottou-98x}. In many applications, including supervised
learning techniques, the function $f$ is the log-likelihood  and $\P$ is an
empirical distribution with density $\frac 1n \sum_{i=1}^n \delta(\sample,\sample_i)$ where
$\{\sample_1,\cdots,\sample_n\}$ is a set of i.i.d. data sampled from an unknown
distribution.

At a given step $t$, SGD can be viewed as a
two-step procedure:
(i) sampling $\sample_t \in \samplespace$ according
to the distribution $\P$; %$\frac{\P}{\P(\samplespace)}$;
(ii) doing an approximate gradient step with step-size $\rho_t$:
\begin{equation}
    \param_{t+1}=\param_{t} - \rho_t \grad_\param f(\sample_t;\param_{t})
\end{equation}
%
%% Note that in many large scale machine learning applications, the number of
%% unique elements in the support of $\P$ is limited, so that the elements are not
%% exactly sampled i.i.d., but explored one by one by doing multiple passes (often
%% called epochs) through the data.

The convergence properties of SGD are directly linked to the
variance of the gradient estimate~\cite{bach2011non}.
Consequently, some improvements on this basic algorithm focus on the use of
(i)~parameter averaging~\cite{Polyak1992} to reduce the variance of the final
estimator, (ii)~the sampling of mini-batches~\cite{Friedlander2011}
when multiple points are sampled at the same time to reduce the variance of the
gradient, and (iii)~the use of adaptive step sizes to have per-dimension
learning rates, e.g., AdaGrad~\cite{duchi2011adaptive}.

In this paper, we propose another general technique, which can be used in
conjunction with the aforementioned ones, which is to \emph{reduce the gradient variance
by learning how to sample training points}.
Rather than learning the fixed optimal sampling distribution and then
optimizing the gradient, we propose to \emph{dynamically learn an optimal
sampling distribution at the same time as the original SGD algorithm}.
Our formulation uses a stochastic process that focuses on the minimization of
the gradient variance, which amounts to do an additional SGD step (to
minimize gradient variance) along each SGD step (to minimize the learning
objective).
There is a constant extra cost to pay at each iteration, but it is the same for
each iteration, and when simulations are expensive or the data access is slow,
this extra computational cost is compensated by the increase of convergence
speed, as quantified in our experiments.

The paper is organized as follows. After reviewing the related work in Section~\ref{sec:relwork},  we show that SGD can be used to find the optimal sampling distribution of an importance sampling estimator (Sec.~\ref{sec:AIS}).
This variance reduction technique is then used during the iterations of a SGD
algorithm by learning how to reduce the variance of the gradient (Sec.~\ref{sec:AWSGD}).
%We then prove that this algorithm converges
%at a rate that depends on the best sampling variance of the gradient~(Sec. \ref{sec:theory}),
We then illustrate this algorithm --- called Adaptive
Weighted SGD (AW-SGD) --- on three well known machine learning problems: image
classification (Sec.~\ref{sec:img}), matrix factorization (Sec.~\ref{sec:mf}),
and reinforcement learning (Sec.~\ref{sec:rl}). Finally, we conclude with a
discussion (Sec.~\ref{sec:conclu}).

\section{Related work}
\label{sec:relwork}
The idea of speeding up learning by modifying the
importance sampling distribution in SGD has been
recently analyzed by~\cite{hazan2011beating} who
showed that a particular choice of the sampling distribution could lead to
sub-linear performance guarantees for support vector machines.
We can see our approach as a generalization of this idea to other models,
by including the learning of the sampling distribution as part of the optimization.
The work of \cite{mineiro2013loss} shows that using a simple model to choose which data to resample from is a useful thing to do, but they do not learn the sampling model while optimizing.
The two approaches mentioned above can be viewed as the extreme case of adaptive sampling, where there
is one step to learn the sampling distribution, and then a second step to learn
the model using this sampling distribution.  The training on language models
has been shown to be faster with adaptive importance
sampling~\cite{senecal2003adaptive, bengio2008adaptive}, but the authors did
not directly minimize the variance of the estimator.

Regarding variance reduction techniques, in addition to the aforementioned ones
(Polyak-Ruppert Averaging~\cite{Polyak1992}, batching~\cite{Friedlander2011},
and adaptive learning rates like AdaGrad~\cite{duchi2011adaptive}), an
additional technique is to use control variates (see for instance
\cite{ross1997simulation}).  It has been recently used by
\cite{paisley2012variational} to estimate non-conjugate potentials in a
variational stochastic gradient algorithm.  The techniques described in this
paper can also be straightforwardly extended to the optimization of a control
variate. A full derivation is given in the appendix, but it was not implemented
in the experimental section. In the neural net community, adapting the order at which the training samples are used is called curriculum learning~\cite{bengio2009learning}, and our approach can be seen under this framework, allthough our algorithm is more general as it can speadup the learning on arbitrary integrals, not only sums of losses over the training data.

Another way to obtain good convergence
properties is to properly scale or rotate the gradient, ideally in the
direction of the inverse Hessian, but this type of second-order method is slow
in practice. However, one can estimate the Hessian greedily, as done in Quasi-Newton
methods such as Limited Memory BFGS, and then adapt it for the SGD algorithm, similarly
to~\cite{Friedlander2011}.

%Finally, note that our Adaptive Weighted SGD method can be coupled with the
%aforementioned variance reduction technique, as we will show in the
%experiments. This integration is straightforward in practice, as we focus on
%making modifications to the sampling part of SGD, which is generally ignored in
%previous works.

%A tradeoff between the
%variance of the gradient and the convergence speed can be obtained using
%batching~\cite{Friedlander2011}. However, both the memory and time
%required for every step increases with the size of the batch.

%The proposed algorithm is centralized on one machine, and an obvious way to use
%multi-core architecture is to distribute the gradient computation.  However,
%for large parameter spaces, the synchronization overhead might limit the
%applicability of the method. Recent work of parallelizing SGD should apply as
%well to the asynchronous distributed setting\cite{agarwal2011distributed}.

%Note that the batching idea (sampling multiple data points at each iteration
%instead of a single one) can also be used to further reduce the variance of
%the gradient.  \paragraph{Convergence proof} An important property of the
%above algorithm is that it indeed minimizes the objective. This has been
%proved by Alexei Gaivoronsky, but is not published yet. The draft of the proof
%is available on request.

%%%%%%%%%%%%%%%%%%%%%%%%%%%%%%%%%%%%%%%%%%%%%%%%%%%%%%%%%%%%%%%%%
\section{Adaptive Importance Sampling}
\label{sec:AIS}

We first show in this section that SGD is a powerful tool to optimize the sampling distribution
of Monte Carlo estimators.  This will motivate our Adaptive Weighted SGD
algorithm in which the sampling distribution is not kept constant, but
learned during the optimization process.

We consider a family $\{\Q_\tau\}$ of sampling distributions on $\samplespace$,
such that $\Q_\tau$ is absolutely continuous with respect to $\P$ (i.e. the support of $\P$ is included in the support of $\Q_\tau$)  for any $\tau$ in the parametric set $\mathcal{T}$.
We denote the density $q = \frac{dQ}{d\P}$. Importance sampling is a common method to estimate the integral in Eq.~\eqref{eq:sgdObjective}. It corresponds to a Monte Carlo estimator of the form (we omit the dependency on $\param$ for clarity):
\begin{eqnarray}
\hat\expf &=& \frac{1}{T} \sum_{t=1}^T \frac{f(\sample_t)}{\q(\sample_t;\tau)},\quad \sample_t\sim \Q_\tau
\label{eq:est1}
\enspace,
\end{eqnarray}
where $\Q_\tau$  is called the \emph{importance distribution}. It is an
\emph{unbiased} estimator of $\expf$, i.e. the expectation of $\hat\expf$ is
exactly the desired quantity $\expf$.

To compare estimators, we can use a variance criterion. The variance of this
estimator depends on $\tau$:
\begin{eqnarray}
\sigma^2(\tau) &=& \Var{\tau}{\hat\expf}
%\nonumber\\
%&=& \frac{1}{T} \Var{\tau}{\frac{f(\sample)}{\q(\sample;\tau)}}
%\nonumber\\
%&=&
=\frac 1T \Expectation{\tau}{\left(\frac{\f(\sample)}{\q(\sample;\tau)}\right)^2} - \frac{\expf^2}{T}
\label{eq:var1}
\end{eqnarray}
% TODO AG removed 1 index of sample (was it necessary?)
%TODO AG: is it true that Expectation{\tau}{f(\sample)} = \expf ?
where $ \Expectation{\tau}{.}$ and $\Var{\tau}{.}$ denote the expectation and
variance with respect to distribution $\Q_\tau$.

To find the best possible sampling distribution in the sampling family
$\{\Q_\tau\}$, one can minimize the variance $\sigma^2(\tau)$ with respect to
$\tau$. If $|f|$ belongs to the family $\{\Q_\tau\}$, then there exists a
parameter $\tau^*\in\mathcal{T}$ such that $q(.,\tau^*)\propto |f|$ $\P$-almost
surely. In such a case, the variance $\sigma(\tau^*)$ of the estimator is null: one can estimate the integral with a single sample.
%
%\mynote{TODO AG}{Add citation for theorem above?}
%
In general, however, the parametric family does not contain a normalized
version of $|f|$. In addition, the minimization of the variance $\sigma^2$ has
often no closed form solution. This motivates the use of approximate
variance reduction methods.

A possible approach is to minimize $\sigma^2$ with respect to the importance
parameter $\tau$. The gradient is:
\begin{eqnarray}
\grad_\tau \sigma^2(\tau)
&=&
\grad_\tau \Expectation{\tau}{\left(\frac{\f(\sample)}{\q(\sample;\tau)}\right)^2}
\label{eq:ExpectationGradientVariance0}
\\
&=&
- \Expectation{\tau}{\frac{\f(\sample)^2\grad_\tau\q(\sample;\tau)}{\q(\sample;\tau)^3}}
\label{eq:ExpectationGradientVariance0}
\\
&=&
-\Expectation{\tau}{\left(\frac{\f(\sample)}{\q(\sample;\tau)}\right)^2 \grad_\tau \log\q(\sample;\tau)}
\nonumber
\enspace.
\end{eqnarray}
This quantity has no closed form solution, but we can use a SGD algorithm with
a gradient step equal on average to this quantity.
To obtain an estimator $g$ of the gradient with expectation given by
Equation~(\ref{eq:ExpectationGradientVariance0}), it is enough to sample a
point $\sample_t$ according to $\Q_\tau$ and then set $g:= -
\f^2(\sample_t)/\q^2(\sample_t;\tau) \grad_\tau \log\q(\sample_t;\tau)$. This is then
repeated until convergence. The full iterative procedure is summarized in
Algorithm~\ref{AIS}.

%%%%%%%%%%%%%%%%%%%%%%%%%%%%%%%%%%%
\begin{algorithm}[t]
\caption{Minimal Variance Importance Sampling}
\label{AIS}
\begin{algorithmic}
\REQUIRE Initial sampling parameter vector $\tau_0\in\TauSpace$
\REQUIRE Learning rates $\{\eta_t\}_{t>0}$
\FOR{$t=0,1,2,\cdots,T-1$}
    \STATE $\sample_t\sim\Q_{\tau_t}$
    \STATE $\tau_{t+1} \gets \tau_{t} + \eta_t \left( \frac{\f(\sample_t)}{\q(\sample_t;\tau_t)}\right)^2 \grad_\tau \log\q(\sample_t;\tau_t)$
\ENDFOR
\STATE Output $\hat\expf \gets \frac{1}{T} \sum_t \frac{f(\sample_t)}{\q(\sample_t;\tau_t)}$
% AG: added time index to last \tau
\end{algorithmic}
\end{algorithm}
%%%%%%%%%%%%%%%%%%%%%%%%%%%%%%%%%%%

In the experiments below, we show that learning the importance weight of an importance sampling estimator using SGD can lead to a significant speed-up in several machine
learning applications, including the estimation of empirical loss functions and
the evaluation of a policy in a reinforcement learning scenario.  In the
following, we show that this idea can also be used in a sequential setting (the
function $f$ can change over time), and when $f$ has multivariate outputs, so
that we can control the variance of the gradient of a standard SGD algorithm
and, ultimately, speedup the convergence.

%\ifa
%\paragraph{Control variates}
%We now describe another variance reduction technique that can also be optimized by gradient descent recently used in a machine context~\cite{paisley2012variational}. The basic idea is that subtracting and then adding a function to $f$ does not change the integral:
%$$
%\gamma = \int_{\OmegaSpace} (f(\omega) - a(\omega;\alpha)) d\P(\omega) + \underbrace{\int_{\OmegaSpace}  a(\omega;\alpha) d\P(\omega)}_{A(\alpha)}
%$$
%where $a:\OmegaSpace\rightarrow\Re\otimes\mathcal{A}$ is called the control variate and we assumed that it can be parameterized by $\alpha$.
%If this function can be efficiently integrated, then we only need to sample the first integral, which is easy in general.
%
%
%%%%%%%%%%%%%%%%%%%%%%%%%%%%%%%%%%%%
%\begin{algorithm}[t]
%\caption{Minimal Variance Importance Sampling}
%\label{AISa}
%\begin{algorithmic}
%\STATE Initialize $\tau_0\in\TauSpace$, $\alpha_0=0$.
%\FOR{$t=1,2,\cdots,T$}
%    \STATE $\omega_t\sim\Q(.;\tau_{t-1})$
%    \STATE $\tau_{t} \gets \tau_{t-1} + \eta_t \left( \frac{\f(\omega_t)-\alpha_t}{\q(\omega_t;\tau_{t-1})}\right)^2 \grad_\tau \log\q(\omega_t;\tau_{t-1})$
%    \STATE (Optional) Update $\alpha_{t}$ using Equation~(\ref{eq:hatAlpha}). Otherwise $\alpha_{t}=\alpha_{t-1}$.
%\ENDFOR
%\STATE Output $\hat\gamma \gets \alpha_t + \frac{1}{T} \sum_{t=1}^T \frac{f(\omega_t)-\alpha_t}{\q(\omega_t;\tau)}$
%\end{algorithmic}
%\end{algorithm}
%%%%%%%%%%%%%%%%%%%%%%%%%%%%%%%%%%%%
%\fi

\section{Biased Sampling in Stochastic Optimization}
\label{sec:AWSGD}

In this section, we first analyze a weighted version of the SGD algorithm where
points are sampled non-uniformly, similarly to importance sampling, and then
derive an adaptive version of this algorithm, where the sampling distribution
evolves with the iterations.

%%%%%%%%%%%%%%%%%%%%%%%%%%%%%%%%%%%%%%%%%%%%%%%%%%%%%%%%%%%%%%%%%%%%%%%%%
\subsection{Weighted stochastic gradient descent}
\label{sec:WSGD}

As introduced previously, our goal is to minimize the expectation of a
parametric function $f$ (cf. Eq.~\eqref{eq:sgdObjective}).
Similarly to importance sampling, we do not need to sample according to the
base distribution $\P$ at each iteration of SGD.
Instead, we can use any distribution $\Q$ defined on $\samplespace$ if each
gradient step is properly re-weighted by the density $q=d\Q/d\P$.
Each iteration $t$ of the algorithm consists in two steps:
(i) sample $\sample_t \in \samplespace$ according to distribution $\Q$;
(ii) do an approximate gradient step:
\begin{equation}
    \param_{t+1} =
    \param_t - \rho_t \frac{\grad_\param f(\sample_t;\param_t)}{q(\sample_t)}  \enspace.
    \label{eq:wsgd}
\end{equation}

Depending on the importance distribution $Q$, this algorithm can have different
convergence properties from the original SGD algorithm.
As mentioned previously, the best sampling distribution would be the one that
gives a small variance to the weighted gradient in Eq.~\eqref{eq:wsgd}. The main
issue is that it depends on the parameters $\param_t$, which are different at
each iteration.

Our main observation is that we can \emph{minimize the variance of the gradient
using the previous iterates}, under the assumption that this variance does not change to quickly when $\param_t$ is updated. We argue this is reasonable in practice as learning rate policies for $\rho_t$ usually assume a small constant learning rate, or a decreasing
schedule~\cite{bottou-98x}.
In the next section, we build on that observation to build a new algorithm that
learns the best sampling distribution $Q$ in an online fashion.

%%%%%%%%%%%%%%%%%%%%%%%%%%%%%%%%%%%%%%%%%%%%%%%%%%%%%%%%%%%%%%%%%%%%%%%%%

\subsection{Adaptive weighted stochastic gradient descent}

Similarly to Section~\ref{sec:AIS}, we consider a family $\{\Q_\tau\}$ of
sampling distributions parameterized by $\tau$ in the parametric set
$\mathcal{T}$. Using the sampling distribution $\Q_\tau$ with p.d.f.
$\q(\sample;\tau)=\frac{d\Q_\tau(\sample)}{d\P(\sample)}$, we can now evaluate the
efficiency of the sampling distributions $Q_\tau$ based on the variance
$\Sigma(\param,\tau)$:
\begin{align}
\Sigma(\param,\tau) :=& \Var{\tau}{\grad_{\param} f(\sample;\param)/\q(\sample,\tau)}
\label{eq:variance0}
\\
=&
\Expectation{\tau}{
\frac{
\grad_{\param} f(\sample;\param)
\grad\transp_{\param} f(\sample;\param)
}
{\q(\sample;\tau)^2}
}
\nonumber \\
~&-\grad_\param \expf(\param)\grad\transp_\param \expf(\param)
\label{eq:variance}
\end{align}
For a given function $f(.;\param)$ we would like to find the parameter
$\tau^*(\param)$ of the sampling distribution that minimizes the trace of the
covariance $\Sigma(\param,\tau)$, i.e.:
\begin{equation}
\tau^*(\param) \in \arg\min_{\tau}
%\trace ( \Sigma(\param,\tau) ) = \arg\min_{\tau}
\Expectation{\tau}{
\left\|\frac{
\grad_{\param} f(\sample;\param)
}
{\q(\sample;\tau)}
\right\|^2
}
\label{eq:minvar}
\end{equation}

%The Equation~(\ref{eq:minvar}) is obtained by remarking that $\Expectation{\tau}{\frac{\grad_{\param} f(\sample_t;\param_{t})}{\q(\param)}}$ does not depend on $Q$. Ideally we would like to compute the best possible proposal $\Q$ at each iteration, i.e. for each different value of $\param_t$.

%QUESTION: Assume that $\f(x,x)$ in strongly convex in $x$ and $\Q_\tau$ in an exponential family distribution.
%Under which assumption on the learning rates $\rho_t$ and $\eta_t$ Algorithm~\ref{ASGD} is guaranteed to converge?
%Intuitively, the usual assumption on $\rho_t$ would be needed ($\sum_t\rho_t=\infty$ and $\sum_t\rho_t^2<\infty$), and the
%convergence relative to $q$ should be less important than the convergence relative to $\x$, meaning that $\eta_t<\rho_t$, but how to prove it?
%

%%%%%%%%%%%%%%%%%%%%%%%%%%%%%%%%%
\begin{algorithm}[t]
\caption{Adaptive Weighted SGD (AW-SGD)}
\label{AWSGD}
\begin{algorithmic}
\REQUIRE Initial target and sampling parameter vectors $\param_0\in\paramspace$ and $\tau_0\in\TauSpace$
\REQUIRE Learning rates $\{\rho_t\}_{t>0}$ and $\{\eta_t\}_{t>0}$
\FOR{$t=0,1,\cdots,T-1$}
    \STATE $\sample_t \sim \Q_{\tau_t}$
	\STATE $d_t \gets \frac{\grad_\param \f(\sample_t;\param_t)}{\q(\sample_t;\tau_t)}$
	\STATE $\param_{t+1} \gets \param_t - \rho_t \d_t$
	\STATE $\tau_{t+1} \gets \tau_t + \eta_t \left\|d_t\right\|^2 \grad_\tau \log\q(\sample_t;\tau_t)$
\ENDFOR
\end{algorithmic}
\end{algorithm}
%%%%%%%%%%%%%%%%%%%%%%%%%%%%%%%%%

\begin{figure*}
	\centering
    \includegraphics[width=1\textwidth]{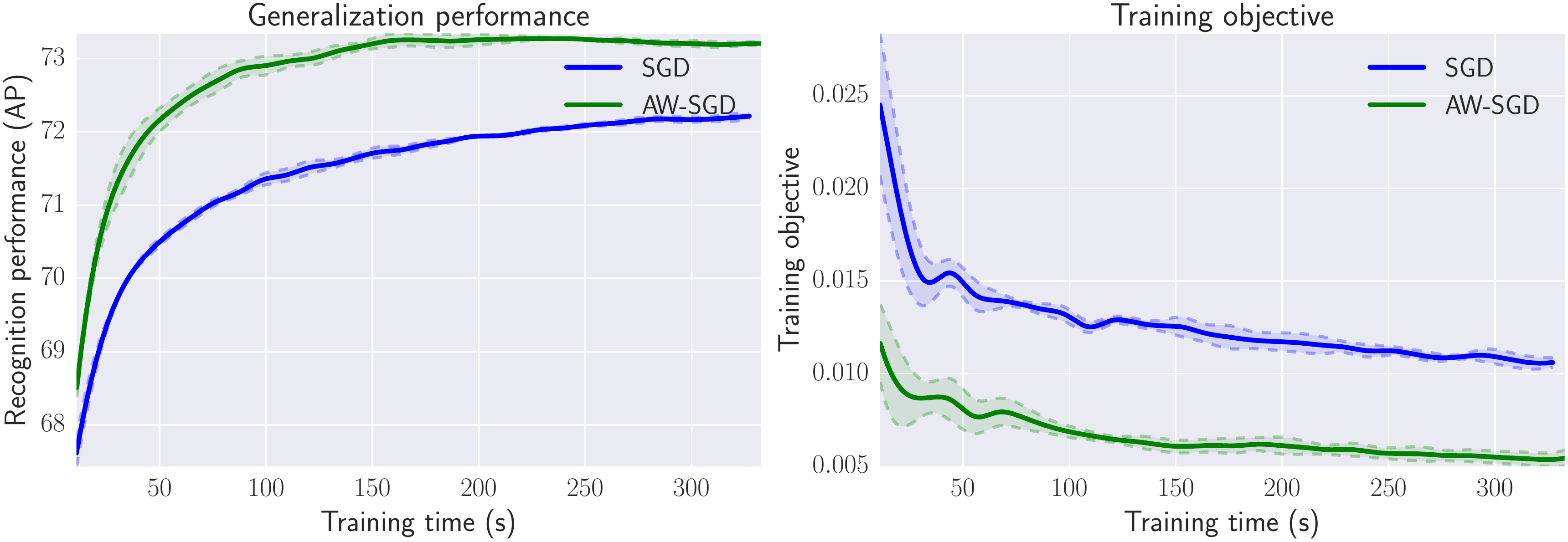}
    % \vspace*{-7mm}
	\caption{Generalization performance (test mean Average Precision) and
training error (average log loss) in function of training time (in seconds)
averaged over three independent runs. SGD converged in $45$ epochs (outside the
graph), whereas AW-SGD converged to the same performance in $10$ times less
epochs for a $982\%$ improvement in training time.
	}
	\label{fig:imgres}
    % \vspace*{-2mm}
\end{figure*}

Note that if the family of sampling distribution $\{Q_\tau\}$ belongs to the exponential family, the
problem~(\ref{eq:minvar}) is convex, and therefore can be solved using (sub-) gradient methods.
Consequently, a simple SGD algorithm with gradient steps having small
variance consists in the following two steps at each iteration $t$:
\begin{enumerate}
    \item perform a weighted stochastic gradient step using distribution
        $Q_{\tau_t}$ to obtain $\param_t$;
    \item compute $\tau_t = \tau^*(\param_{t})$ by solving
        Equation~(\ref{eq:minvar}), i.e. find the parameter $\tau_t$ minimizing
        the variance of the gradient at point $\param_t$. This can be done approximately by applying $M$ steps of stochastic gradient descent.
\end{enumerate}
The inner-loop SGD algorithm involved in the second step can be based on the current sample, and
the stochastic gradient direction is
\begin{align}
\grad_\tau \Sigma(\param_t,\tau)
=&
\grad_\tau \Expectation{\tau}{
    \left\|
    \frac{\grad_{\param_t}\f(\sample;\param_t)}{\q(\sample;\tau)}
    \right\|^2}
\label{eq:GradExpectationGradientVariance}
\\
=&
-\Expectation{\tau}{
    \left\|
    \frac{\grad_{\param_t} \f(\sample;\param_t)}{\q(\sample;\tau)}
    \right\|^2
    \grad_\tau \log\q(\sample;\tau)}
\nonumber %\enspace.
\end{align}
%We can obtain an unbiased estimator of this gradient by first sampling a point $\sample_t$ using
%$Q_{\tau_t}$ and then computing the vector
In practice, we noted that it is enough to do a single step of the inner loop, i.e. $M=1$.
We call this simplified algorithm the \emph{Adapted-Weighted SGD Algorithm} and its pseudocode is
given in Algorithm~\ref{AWSGD}.
We see that AW-SGD is a slight modification of the standard SGD --- or any variant of it, such as Adagrad, AdaDelta or RMSProp -- but where the sampling distribution evolves during the algorithm, thanks to the update of $\tau_t$. This algorithm is useful when the gradient has a variance that can be significantly reduced by choosing better samples.
An important design choice of the algorithm is the choice of the decay of the step sizes sequences $\{\rho_t\}_{t>0}$ and $\{\eta_t\}_{t>0}$. While using adaptive step sizes appears to be useful in some settings, it appears that the regime in which AW-SGD outperforms SGD is when $\eta_t$ are significantly larger than $\rho_t$, meaning that the algorithm converges quickly to the smallest variance, and AW-SGD tracks it during the course of the iterations. Ideally, the sequence of sampling parameters $\{\tau_t\}$ remains close to the optimal trajectory which consist is the best possible sequence of sampling parameters given by Equation~\ref{eq:minvar}
%
% \begin{eqnarray}
% \tau^*_t :=\arg\min_\tau \Var{\Q_{\tau}}{\frac{\grad\f_{w_t}(x_t)}{\q_\tau(x_t)}}
% \enspace.
% \end{eqnarray}

%

%We prove in the additional material
%that under regularity conditions on $f$ and $\{Q_\tau\}$ and sufficiently large
%decreasing learning rates $\{\rho_t\}$ and $\{\eta_t\}$, AW-SGD
%converges in probability to a local minimum of $f$. The proof is based on the
%control of the error due to the non-constant learning rate.  It uses
%results on stochastic optimization from \cite{gaivoronskii1978nonstationary, pflug1996optimization}.

%
% \begin{figure}[t]
% 	\centering
%     \includegraphics[width=\linewidth]{awsgd_image_sampling_timings.pdf}
%     \caption{Timings of the different operations involved at each training
%         step of our image classifier. The additional per-update cost of AW-SGD
%         is minimal compared to SGD (increase of $1.7\%$ of the total compute time),
%         and significantly compensated for by the acceleration in the
%         convergence as shown in our experiments. All experiments were run on an
%         Intel(R) Xeon(R) E5-2630L CPU with a Nvidia Tesla K40m GPU.
% 	}
% 	\label{fig:imgtimings}
% \end{figure}

%\vspace{1cm}

%%%%%%%%%%%%%%%%%%%%%%%%%%%%%%%%%%%%%%%%%%%%%%%
We now illustrate the benefit of this algorithm in three different applications: image classification, matrix factorization and reinforcement learning.

\section{Application to Image classification}
\label{sec:img}
\newcommand{\data}{\mathcal{D}}
\newcommand{\trans}{\phi}
\newcommand{\tparam}{\theta}
\newcommand{\tparamspace}{\Theta}
\newcommand{\xt}{\trans_{\tparam}(I_{i_t})}
\newcommand{\yt}{y_{i_t}}
\newcommand{\ei}{\mathbf{e_{i_t}}}

Large scale image classification is an important machine learning tasks where images containing a given category are much less frequent than images not containing the category. In practice, to learn efficient classifiers, one need to optimize a class-imbalance hyper-parameter~\cite{Perronnin2012}.
Furthermore, as suggested by the standard practice of ``hard negative
mining''~\cite{Dalal2005}, positives and negatives should have a different importance during optimization, with positives being more important at first, and negatives gradually gaining in importance.
However, cross-validating the best imbalance hyper-parameter at each iteration is prohibitively expensive.
Instead, we show here that AW-SGD can be used for biased sampling depending on the label, where the bias $\tau_t$ (imbalance factor) is adapted along the learning.
%
%% As AW-SGD converges in less steps than standard SGD, we need to sample fewer
%% images, and, therefore, extract less feature vectors. Note that the additional
%% fixed cost of AW-SGD compared to SGD is negligible with respect to the feature
%% extraction operation, the main bottleneck in image classification with modern
%% high-level features (see Figure~\ref{fig:imgtimings} for precise timings in our
%% case). Therefore, it takes both less steps \emph{and} less time to reach a
%% target performance with AW-SGD.

To measure the acceleration of convergence, we experiment on the widely used
Pascal VOC 2007 image classification benchmark~\cite{Everingham2010}. Following
standard practice~\cite{Donahue2014, Oquab2014, Girshick2014}, we learn a
One-versus-Rest logistic regression classifier using deep image features from
the last layers of the pre-trained AlexNet Convolutional
Network~\cite{Krizhevsky2012}.
Note that this image classification pipeline provides a strong baseline,
comparable to the state of the art~\cite{Oquab2014}.

Let $\mathcal{D}=\{(I_i, y_i), i=1, \cdots, n\}$ a training set of $n$
images $I_i$ with labels $y_i \in \{-1, 1\}$.
The discrete distribution over samples is parametrized by the log-odd $\tau$
of the probability of sampling a positive image: the family of sampling distributions $\{Q_\tau\}$ over
$\mathcal{D}$ can be written as:
\begin{flalign}
\q(\sample; \tau)
=& \frac{n}{n(\y_i)} \varsigma(\y_i \tau)
\label{eq:VSq1}
\end{flalign}
with $\varsigma(a):=1/(1+e^{-a})$ representing the sigmoid function\footnote{Using the sigmoid link enables an optimization in the real line instead on the constrained set $[0,1]$}, $\sample=i$, an image index in  $\{1, \dots, n\}$, $\tau \in \Re$, and
$n(+1)$ (resp. $n(-1)$) is the number of positive (resp. negative) images.
With this formulation, the update equations in AW-SGD (Algo.~\ref{AWSGD}) are:
\begin{eqnarray}
\f(\sample_t;\param_t)
&=& \ell\left( f(\xt;\param_t), \yt \right)
\nonumber \\
&=& \log\left(1 + \exp(-\yt \param_t^T \xt)\right)
\label{eq:VSlogloss} \\
&=& -\log\left(s_t\right)
\label{eq:VSst}
\end{eqnarray}
with $s_t:=\varsigma(\yt \param_t^T \xt)$ representing the predicted probability and $\theta$ the parameters of the feature function.
\begin{eqnarray}
\grad_\param \f(\sample_t;\param_t)
&=& \left(s_t - 1 \right) \yt \xt
, \nonumber\\
\nabla_\tau\log\q(\sample_t;\tau_t) &=& \yt (1 - s(\yt \tau_t)) \ .
\label{eq:gradLogQImg}
\end{eqnarray}

We initialize the positive sampling bias parameter with the value $\tau_0=0.0$,
which yields a good performance both for SGD and AW-SGD.
For both the SGD baseline and our AW-SGD algorithm we use AdaGrad~\cite{duchi2011adaptive} to choose the learning rates $\rho_t$ and $\eta_t$. Both were initialized at 0.1.

Figure~\ref{fig:imgres} shows that AW-SGD converges faster than SGD for both
training error and generalization performance. Acceleration is both in time and
in iterations, and AW-SGD only costs $+1.7\%$ per iteration with respect to SGD in our
implementation.
In further experiments, we noticed that the positive sampling bias parameter $\tau_t$ indeed gradually decreases, i.e. the /{algorithm learns that it should focus more on the
harder negative class. We also show that the values learned for this sampling
parameter also depend on the category.

\begin{figure}[H]
	\centering
    \includegraphics[width=1.0\linewidth]{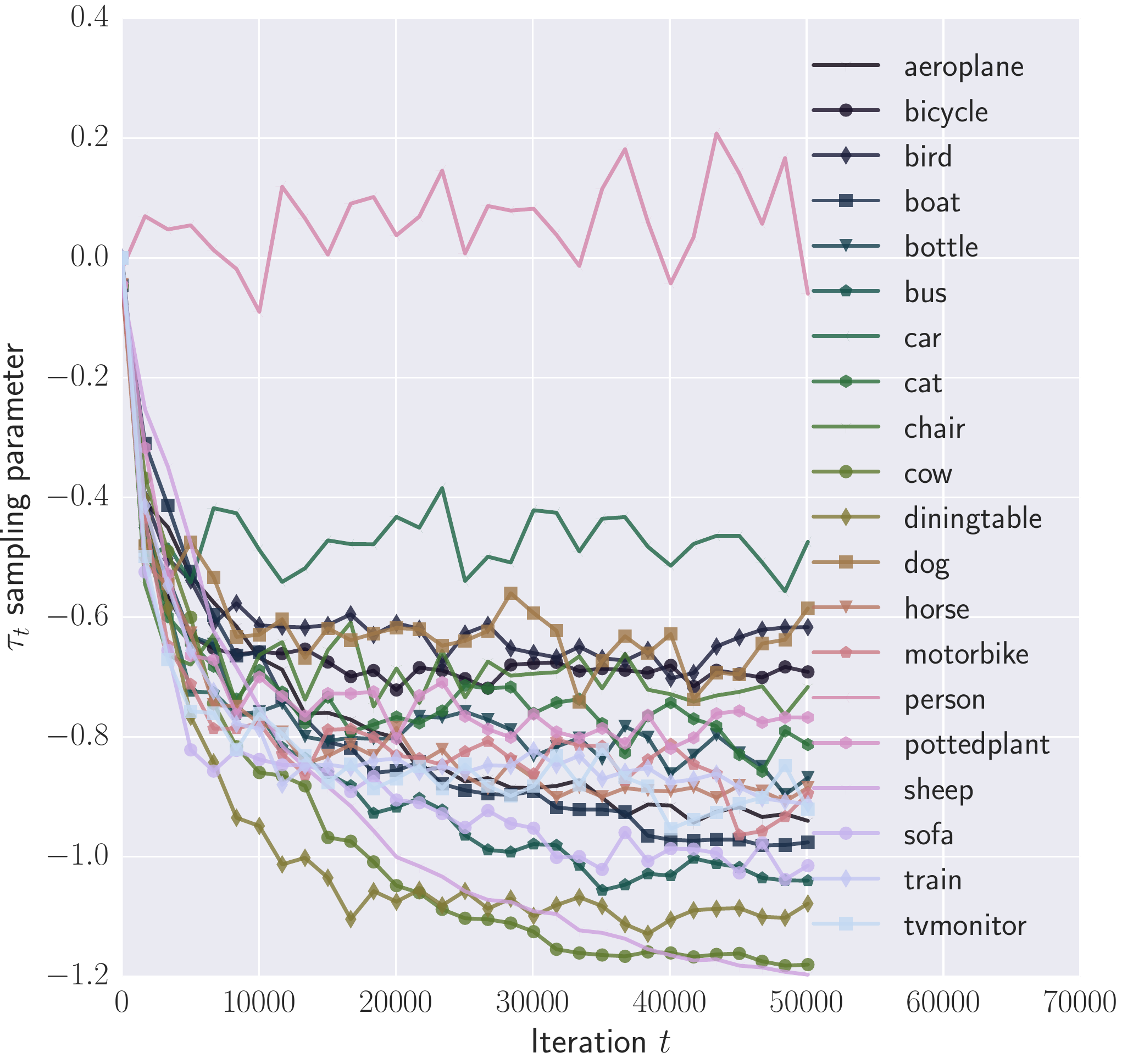}
    % \vspace*{-5mm}
    \caption{Evolution of the positive sampling bias parameter $\tau_t$ in
        function of the training iteration $t$ for the different object
        categories of Pascal VOC 2007.
	}
	\label{fig:imgtaus}
\end{figure}
Figure~\ref{fig:imgtaus} displays the evolution of the positive sampling bias
parameter $\tau_t$ along AW-SGD iterations $t$. Almost all classes expose the
expected behavior of sampling more and more negatives as the optimization
progresses, as the negatives correspond to anything but the object of interest,
and are, therefore, much more varied and difficult to model.
The ``person'' class is the only exception, because it is, by far, the category
with the largest number of positives and intra-class variation.
Note that, although the dynamics of the $\tau_t$ stochastic process are
similar, the exact values obtained vary significantly depending on the class,
which shows the self-tuning capacity of AW-SGD.

\section{Application to Matrix factorization}
\label{sec:mf}

We applied AW-SGD to learn how to sample
the rows and columns in a SGD-based low-rank matrix decomposition algorithm.
Let $Y\in\Re^{n\times m}$ be a matrix that has
been generated by a rank-$K$ matrix $UV\transp$, where $U\in\Re^{n\times K}$
and $V\in\Re^{m\times K}$.
%To handle matrices with continuous, counts or binary values,
We consider a differentiable loss function $\ell(z;y)$ where $z\in\Re$
and $y$ is the observed value. With the squared loss,
we have each entry of $Y$ is a real scalar and $\ell(z,y)=(z-y)^2$. The full loss function is
\begin{eqnarray}
\expf(U,V) &=& \sum_{i=1}^n\sum_{j=1}^m  \ell(u_i v_j\transp,y_{ij})
\label{eq:mfobjective}
\end{eqnarray}
We consider the  sampling distributions $\{Q_\tau\}$ over the set
$\samplespace := \{1, \cdots, n\} \times \{1, \cdots, m\}$,
where we independently sample a row $i$ and a column $j$ according
to the discrete distributions $\softmax(\tau')$ and $\softmax(\tau'')$
respectively, with $\tau'\in\Re^n$, $\tau'' \in \Re^m$, $\tau = (\tau',\tau'') \in \Re^{m+n}$,
and $\sample = (i,j)$. We define:
\begin{flalign}
	\softmax(\mathbf{z}) =& (e^{z_1},e^{z_2},\cdots,e^{z_p})/\left(\sum_{i=1}^p e^{z_i}\right)
	\label{eq:MFsm}
	~\\
	\q(x,\tau) =& \softmax(\tau') \softmax(\tau'')
	\label{eq:MFq}
\end{flalign}
with $\softmax: \Re^p \mapsto \Re^p$ the softmax function.
Using the square loss, as in the experiments below, the update equations in AW-SGD (Algo.~\ref{AWSGD}) are:
\begin{flalign}
	\f(\sample_t; u_t, v_t)
	=& \ell(u_{i_t} v_{j_t}\transp,y_{i_t j_t})
%	\nonumber \\
	= (u_{i_t} v_{j_t}\transp - y_{i_t j_t})^2
%	\label{eq:MFsquareloss} \\
	= s_t^2
%	\label{eq:MFsquarelossdt} \\
\end{flalign}
\begin{flalign}
	\grad_{u_{i_t}} \f(\sample_t; u_t, v_t)
	= 2v_{j_t} s_t,\\%\quad \quad
	\grad_{v_{j_t}} \f(\sample_t; u_t, v_t)
	= 2u_{i_t} s_t
	\label{eq:MFgradsquaredloss}
\end{flalign}
\begin{flalign}
	\nabla_{\tau'}\log\q(\sample_t;\tau_t) = e_i - \softmax(\tau'),\\%\quad \quad
	\nabla_{\tau''}\log\q(\sample_t;\tau_t) = e_j - \softmax(\tau'')
	\label{eq:gradLogQMF}
\end{flalign}
where $e_i \in \Re^n$ and $e_j \in \Re^m$, vectors with $1$ at index $i$ and $j$ respectively,
and all other components are $0$.

%An alternative distribution is to assume that we know a set of $C$ positive matrices $M^{(1)},\cdots, M^{(C)}$ the same size as $Y$
%and which sum to one. The distribution $Q_\tau$ is assumed to be a mixture distribution where $\tau_1,\cdots,\tau_K$ are the parameters of the component prior
%probabilities and the matrices $M^{(c)}$, $c=1,\cdots,L$ are the component distributions:
%\begin{eqnarray}
%\q(i,j) &=& \sum_{c=1}^C \frac{e^{\tau_c}}{\sum_{c'=1}^C e^{\tau_{c'}}} M_{ij}^{(c)}
%\label{eq:mfQ1}
%\end{eqnarray}
%The required derivative is the following:
%\begin{eqnarray}
%\grad_{\tau} \log \q(i,j) &=&  \left(
%\frac{M_{ij}^{(c)}e^{\tau_c}}{\sum_{c'=1}^C M_{ij}^{(c)} e^{\tau_{c'}}}
%-
%\frac{e^{\tau_c}}{\sum_{c'=1}^C e^{\tau_{c'}}}
%\right)_{c=1}^C
%\label{eq:mfQ1}
%\end{eqnarray}

\begin{figure}
	\centering
    \includegraphics[width=0.99\linewidth]{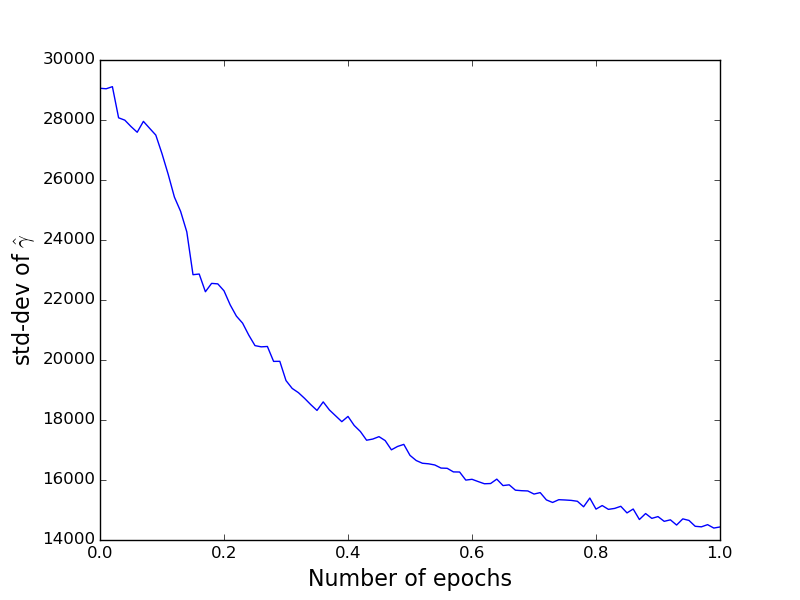}\\
    \caption{Results of the Minimal Variance Importance Sampling algorithm (Algorithm 1.). The curve shows the standard deviation of the estimator of the loss $\hat\gamma$ as a function of the number of matrix entries that have been observed.
	}
	\label{fig:mf1}
\end{figure}

In the matrix factorization experiments, we used the mini-batch technique with batches of size 100, $\rho_0$ and $\eta_0$ were tuned to yield the minimum $\expf$ at convergence, separately with each algorithm. All results are averaged over 10 runs.
$\tau'$ and $\tau''$ were initialized with zeros to get an initial uniform sampling
distribution over the rows and columns. The model learning rate decrease was set to $\rho_0/((N/2)+t)$, $\eta_0$ was kept constant.

%Hierarchical softmax?

In Figure~\ref{fig:mf1}, we simulated a $n\times m$ rank-$K$ matrix, for $n=m=100$ and
$K=10$, by sampling $U$ and $V$ using independent centered Gaussian variables with unit variance.
To illustrate the benefit of adaptive sampling, we multiply by 100 a randomly drawn square block of size 20, to experimentally
observe the benefit of a non uniform sampling strategy.
The results of the minimal variance important sampling
scheme (Algorithm~\ref{AIS}) is shown on the left.
%At each iteration, we compute the exact variance of the IS estimator
%$\hat\expf$ on the full matrix.
We see that after having seen 50\% of the
number $N=nm$ of matrix entries, the standard deviation of the importance sampling estimator is divided by two, meaning that we would need only half of the samples to evaluate
the full loss compared to uniform sampling .
Figure~\ref{fig:mf2} shows the loss decrease of
SGD and AW-SGD and on the same matrix for multiple learning rates.
The $x-$axis is expressed in epochs, where one epoch corresponds to $N$ sampling of values in the matrix.
AW-SGD converges significantly faster than the best uniformly sampled SGD, even after 1 epoch through the data. On
average, AW-SGD requires half of the number of iterations to converge to the same value.

\begin{figure}[t]
	\centering
    \includegraphics[width=0.99\linewidth]{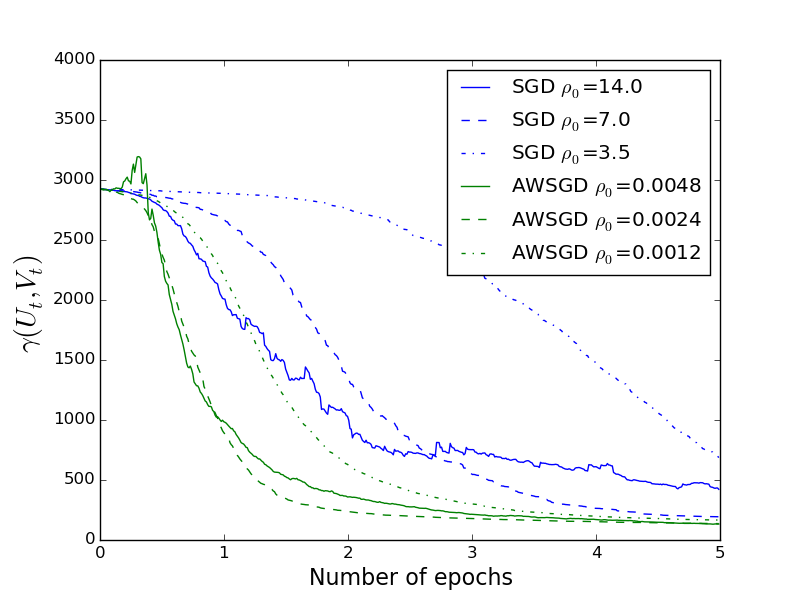}
    \caption{Comparison
        of the convergence speed of the AW-SGD algorithm compared
        to the standard SGD algorithm (uniform sampling of rows and columns) on the matrix factorization experiment.
	}
	\label{fig:mf2}
\end{figure}

%\begin{figure*}[t]
%%\center
%\centering
%\includegraphics[width=0.33\linewidth]{plot_mnist.png}
%\includegraphics[width=0.33\linewidth]{plot_antenna.png}
%\includegraphics[width=0.33\linewidth]{plot_conll.png}
%\caption{Evolution of the training error as a function of the number of epochs
%    for the uniformly-sampled SGD and the Adaptive Weighted
%    SGD, using the best $\rho_0$ for each algorithm.\label{fig:realdata_mf}}
%\end{figure*}

\paragraph{In-painting experiment} We compared both algorithms on the MNIST dataset \cite{lecun1998gradient}, on which low-rank
decomposition techniques have been successfully applied \cite{das2011non}.
We factorized with $K=50$ the training set for the zero digit, a $5923\times784$ matrix,
where each line is a $28\times28$ image of a handwritten zero, and each column one pixel.
Figure ~\ref{fig:mnist1} shows the loss decrease for both algorithms on the first iteration.
AW-SGD requires significantly less samples to reach the same error.
At convergence, AW-SGD showed an average $2.52\times$ speedup in execution time compared to SGD, showing that
its sampling choices compensate for its parametrization overhead.

\paragraph{Non-stationary data} On Figure~\ref{fig:mnist2} we progressively substituted images of handwritten zeros by images of handwritten ones.
It shows, every 2000 samples (i.e. 0.0005 epoch), the heatmap of the sampling probability of each pixel, $\varsigma (\tau'') $,
reorganized as $28\times28$ grids. Substitution from zeros to ones was made between 10000 and 20000 samples (on 2nd line).
One can distinctly recognize the zero digit first, that progressively
fades out for the one digit\footnote{We created an animated gif with more of these images and inserted it in the supplementary material.}.
This transitions shows that AW-SGD learns to sample the digits that are likely to have  a high impact on the loss. The algorithm adapts online to changes in the underlying distribution (transitions from one digit to another).

%\begin{itemize}
%   \item \textbf{mnist}: a 100x784 matrix containing hand-written zero digits
%       images in rows, and the corresponding grey level of each pixel (from 0
%       to 255) in columns.
%   \item \textbf{antenna}: This dataset consist of (anonymized) mobile phone
%       connectivity data of foreign tourists traveling within a large European
%       city. A typical day of the week was split in five-minute intervals, and
%       the number of connections to each antenna (or base station) in this time
%       period was computed, indicating which locations are preferred by
%       tourists like at what time of the day. This results in a 723x288 matrix.
%   \item \textbf{conll}: A dataset extracted from a annotated corpus of texts.
%       Where each entry of the matrix corresponds to carefully chosen word
%       types in one of 19 pre-selected categories, to form a 2207x19 matrix.
%\end{itemize}
%We set $ K = 50$ for mnist and antenna and $K=10$ for conll for the factor
%matrices, the figure \ref{fig:realdata_mf} shows the evolution of
%$\expf$ for each algorithm with their respective optimal
%$\rho_0$ and $\eta_0$. In every case, we observe a significant reduction of
%the error when we use AW-SGD instead of the uniformly sampled SGD.
\begin{figure}[t]
	\centering
    \includegraphics[width=1\linewidth]{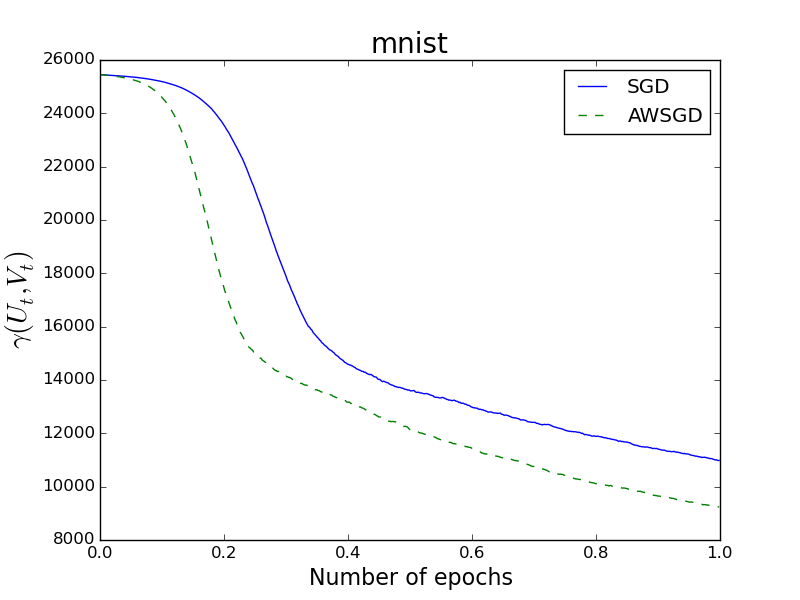}\\
    \vspace*{-5mm}
    \caption{Evolution of the training error as a function of the number of epochs
		for the uniformly-sampled SGD and AW-SGD for the matrix factorization application applied on MNIST data.
	}
	\label{fig:mnist1}
    \vspace*{-4mm}
\end{figure}

Combined with adaptive step size algorithms such as AdaGrad, we noticed that Adagrad did not
improved the convergence speed of AW-SGD in our matrix factorization experiments.
A possible explanation is that the adaptive sampling favors some
rows and columns, and AdaGrad compensates the non-uniform sampling,
such that using AW-SGD and AdaGrad simultaneously converges only slightly faster than AdaGrad alone.
It should behave similarly on other parameterizations of $\tau$ where $\tau$ indices are linked to parameters indices.
However, in many of our experiments, Adagrad performances were not matching the best cross-validated learning rates.

\section{Sequential Control through Policy Gradient}
\label{sec:rl}

%\begin{figure}[t]
%\includegraphics[width=0.5\linewidth]{rl_figure_1.png}
%
%\begin{tabular}{|l|l|l|}
%   \hline
%    $\ell$ & AW-SGD & SGD\\
%   \hline
%    15   & $0.91\pm0.021$ &  $0.85\pm0.032$   \\
%    50   & $0.85\pm0.031$ &  $0.77\pm0.042$   \\
%    80  & $0.81\pm0.046$ &  $0.74\pm0.056$  \\
%   \hline
%\end{tabular}
%
%\caption{Top: Cumulated sum of success obtained by SGD and AW-SGD policy
%    iteration on a squared grid world with $l = 15$. Bottom: Probability of
%    success, e.g. reaching the end point, for various environment sizes.
%    \label{fig:xp1-all}
%}
%\label{fig:rl-xp-all}
%\end{figure}
%%%%%%%%%%%%%%%%%%%%%%%%%%%%%%%%%%%%%%%%%%%%%%%%%%%

Stochastic optimization is currently one of the most popular approaches for policy learning in the context of Markov Decision Processes. More precisely, policy gradient has
become the method of choice in a large number of contexts in reinforcement learning ~\cite{SilverLHDWR14,Munos2006}. Here, optimizing
the integral (1) is related to policy gradient algorithms which aim at
minimizing an expected loss (i.e. a negative reward or a cost) or maximizing a reward in an episodic setting (i.e. with a predefined finite trajectory length) and off-policy estimation.
Equivalently, if we consider the sampling space as being the (action, state) trajectory of a Markov Decision Process, AW-SGD can be viewed as a off-policy gradient algorithm, where $P_\w$ and $Q_\tau$ have the same parameterization, i.e. $\mathcal{W}=\mathcal{T}$. The objective is to maximize the expected reward for the target policy $P\w$, and to minimize the variance of the gradient for the policy gradient for the exploration policy $\Q_\tau$.

We considered a canonical grid-world problem \cite{SUTTON92A,szepesvari2010algorithms} with a squared grid of size $\ell$ is considered. A classical reward setting has been applied: the reward function is a discounted instantaneous reward of $-1$ assigned on each cell of the grid and a reward of $1000$ for a terminal state located at the down right of the grid.
In this context, an episode is considered as successful if the defined terminal state is reached.  Finally, a random distribution of $n_{\mathrm{trap}} = \frac{\ell}{25}$ terminal states with a negative reward of $-1000$ are also positioned. The start state is located at the very up-left cell of the grid.

In this experimental setting, the parameters $\param$ and $\tau$ of the target policy $\P_\param$ and the exploration policy $Q_\tau$ are defined in the space $\Re^{\ell\times\ell \times4}$. More precisely, the probability of an action $a$ at each position $\{x,y\} \in [1,\ell]^2$ follows a multinomial distribution of parameters $\{p^{x,y}_1,\ldots,p^{x,y}_4\}$. Indeed,
in the context of the grid type of environment that we will use in this section, these parameters basically correspond to the log-odds of the probability of moving in one of the four directions at each position of the grid (movements outside the grid do not change the position). The distribution $\Q_\tau$ of sampled trajectories are different from the distribution of trajectory derived from $P_\param$ (off-policy learning).

%%%%%%%%%%%%%%%%%%%%%%%%%%%%%%%%%%%%%%%%%%%%%%%%%%%
\begin{figure}[t]
	\centering
	\includegraphics[width=1\linewidth]{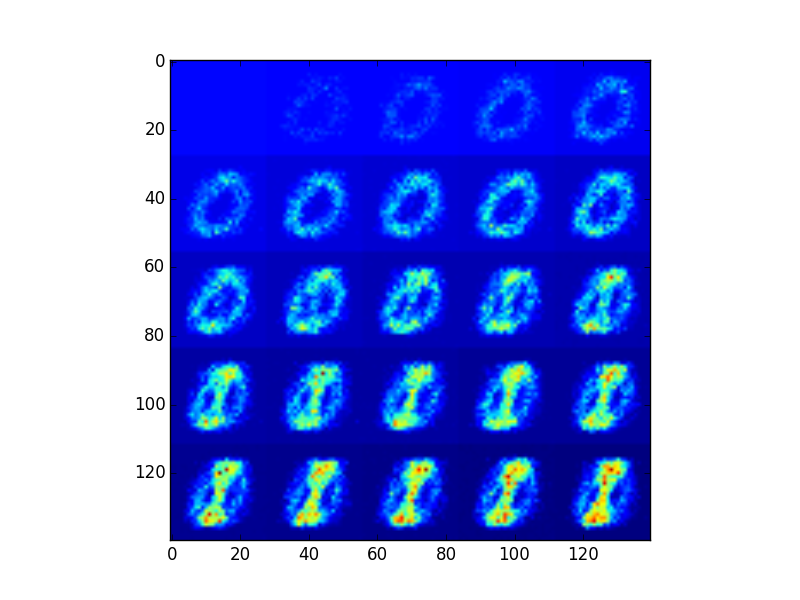}
    \vspace*{-5mm}
    \caption{Illustration of the evolution of the sampling distribution when data are not i.i.d. Each heatmap contains the sampling probability of each pixel in the MNIST matrix factorization experiment.
	}
	\label{fig:mnist2}
    \vspace*{4mm}
\end{figure}

%%%%%%%%%%%%%%%%%%%%%%%%%%%%%%%%%%%%%%%%%%%%%%%%%%%
The policy is optimized using Algorithm~\ref{AWSGD}.  The baseline corresponds
to a policy iteration based on SGD where trajectories are sampled using their
current policy estimate (on-policy learning). On
Table~\ref{tab:xp1-all}, the table gives the average means and variances
obtained for a batch of $20000$ learning trials using both algorithms with
properly tuned learning rate (the optimal learning rate is different in the two
algorithms, for SGD $\rho=2.1$ and for AWSGD $\rho=0.003$ has been found). We
can see that for all the tested grid sizes, there is a significant improvement
(close to 10\% relative improvement) of the expected success when adaptive
weighted SGD is used instead of the on-policy learning SGD algorithm.

%On the
%top of Figure~\ref{fig:rl-xp-all}, we observe an improvement of the
%converge speed of the AWSGD compared to the on-policy learning SGD algorithm.

\section{Adapting to Non-Uniform Architectures}

In many large scale infrastructures, such as computation servers shared by many people, data access or gradient computation are unknown in advance. For example, in large scale image classification, some images might be stored in the RAM, leading to a access time that is order of times faster than other images stored only on the hard drive.
These hardware systems are sometimes called Non-Uniform Memory Access~\cite{nieplocha1996global}. It is also the case in matrix factorization, when some embeddings are stored locally, and others downloaded through the network. How can we inform the algorithm that it should sample more often data stored locally?

A simple modification of AW-SGD can enable the algorithm to adapt to non-uniform computation time. The key idea is to learn dynamically to minimize the expected loss decrease per unit of time. To make AW-SGD take into account this access time, we simply weighted the update of $\tau$ in Algorithm~\ref{AWSGD} by
dividing it by the simulated access time $\Delta_x$ to the sample $x$. This is summarized in Algorithm~\ref{alg:hardware}.

As an experiment, we show in the matrix factorization case that the time-aware AW-SGD is able to learn and exploit the underlying hardware when the data does not fit entirely in memory, and one part of them has an extra access cost. To do so, we generate a $1000\times 1000$ rank-$10$ matrix, but without high variance block, so that variance is uniform across rows and columns. For the first half of the rows of the matrix, i.e. $i<\frac{n}{2}$, we consider the data as being in main memory, and simulate an access cost of 100ns for each sampling in those rows, inspired by Jeff Dean's explanations\cite{DeanLatency}. For the other half of the rows, $i >= \frac{n}{2}$, we multiply that access cost by a factor $f$,
we'll call the slow block access factor.
The simulated access time to the sample $(i,j)$, $\Delta_{(i,j)}$ is thus given by:
$\Delta_{(i,j)} = 10^{-7} \times f $ if $i >= \frac{n}{2}$, and $\Delta_{(i,j)} = 10^{-7}$ if $i >= \frac{n}{2}$. We ranged the factor $f$ from 2 to $2^{20}$.

\begin{table}[t]
	\centering
	\begin{tabular}{|l|l|l|}
	   \hline
		$\ell$ & AW-SGD & SGD\\
	   \hline
		15   & $0.91\pm0.021$ &  $0.85\pm0.032$   \\
		50   & $0.85\pm0.031$ &  $0.77\pm0.042$   \\
		80  & $0.81\pm0.046$ &  $0.74\pm0.056$  \\
	   \hline
	\end{tabular}
	\caption{Probability of success, e.g. reaching the end point, for various environment sizes.
	\label{tab:xp1-all}}
\end{table}

The time speedup achieved by the time-aware AW-SGD against SGD is plotted against the evolution of this factor in Figure~\ref{fig:speedups}. For each algorithm, we summed the real execution time and the simulated access times in order to take into account the time-aware AW-SGD sampling overhead.
The speedup is computed after one epoch, by dividing SGD total time
by AW-SGD total time. Positive speedups starts with a slow access time factor $f$ of roughly 200, which corresponds to a random read on a SSD. Below AW-SGD is slower, since the data is homogeneous, and time access difference is not yet big enough to compensate its overhead.
At $f=5000$, corresponding to a read from another computer's memory on local network,
speedup reaches $10\times$. At $f=50000$, a hard drive seek, AW-SGD is $100\times$ faster.
This shows that the time-aware AW-SGD overhead is compensated by its sampling choices.

Figure~\ref{fig:numa_loss} shows the loss decrease of both algorithms on the 5 first epochs with $f=5000$. It shows that  if the access time was the uniform, AW-SGD would have the same convergence speed as standard SGD (this is expected by the design of this experiment). Hence, even in such case where there is no theoretical benefit of using the time-aware AW-SGD in terms of epochs, the fact that we learn the underlying access time to bias the sampling could potentially lead to huge improvements of the convergence time.

%%%%%%%%%%%%%%%%%%%%%%%%%%%%%%%%%
\begin{algorithm}[t]
\caption{Time-Aware AW-SGD)}
\label{alg:hardware}
\begin{algorithmic}
\REQUIRE Initial values for $\param_0\in\paramspace$ and $\tau_0\in\TauSpace$
\REQUIRE Learning rates $\{\rho_t\}_{t>0}$ and $\{\eta_t\}_{t>0}$
\FOR{$t=0,1,\cdots,T-1$}
  \STATE $s_t \gets \mathrm{getCurrentTime()}$
  \STATE $\sample_t \sim \Q_{\tau_t}$
	\STATE $d_t \gets \frac{\grad_\param \f(\sample_t;\param_t)}{\q(\sample_t;\tau_t)}$
	\STATE $\param_{t+1} \gets \param_t - \rho_t \d_t$
  \STATE $e_t \gets \mathrm{getCurrentTime()}$
	\STATE $\tau_{t+1} \gets \tau_t + \frac{\eta_t}{e_t - s_t} \left\|d_t\right\|^2 \grad_\tau \log\q(\sample_t;\tau_t)$
\ENDFOR
\end{algorithmic}
\end{algorithm}
%%%%%%%%%%%%%%%%%%%%%%%%%%%%%%%%%

\section{Conclusion}
\label{sec:conclu}

In this work, we argue that SGD and importance sampling can strongly benefit from each other.
SGD algorithms can be used to learn the minimal variance sampling distribution,
while importance sampling techniques can be used to improve the gradient estimation of SGD
algorithm. We have introduced a simple yet efficient Adaptive Weighted SGD
algorithm that can optimize a function while optimizing the way it samples the
examples. We showed that this framework can be used in a large variety of
problems, and experimented with it in three domains that have apparently no direct
connections: image classification, matrix factorization and reinforcement learning.
In all the cases, we
can gain a significant speed-up by optimizing the way the samples are
generated.

There are many more applications in which these variance reduction
techniques have a strong potential. For example, in variational inference, the objective
function is an integral and SGD algorithms are often used to increase
convergence~\cite{hoffman2013stochastic, paisley2012variational}. Computing these integrals stochastically could be made more efficient by sampling non-uniformly in the integration space.
Also,
the estimation of intractable log-partition function, such as Boltzmann
machines, are potential candidate models in which importance sampling has
already been proposed, but without variance reduction
technique~\cite{salakhutdinov2008quantitative}.

This work also shows that we can learn about the algorithm while optimizing, as shown by the time-aware AW-SGD. This idea can be extended to design new types of meta-algorithms that \emph{learn to optimize} or \emph{learn to coach other algorithms}.

\begin{figure}[t]
	\centering
    \includegraphics[width=1\linewidth]{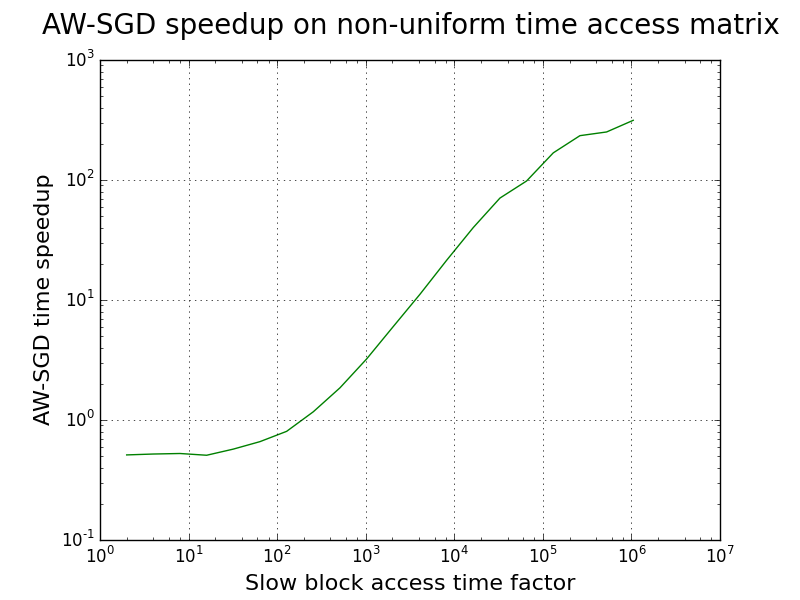}
    \caption{ Evolution of the training error as a function of the number of epochs
			on the simulated matrix with different access costs, with $f=5000$,
			for the uniformly-sampled SGD and AW-SGD using best $\rho_0$ for each algorithm.
	}
	\label{fig:speedups}
\end{figure}

\begin{figure}[t]
	\centering
    \includegraphics[width=1\linewidth]{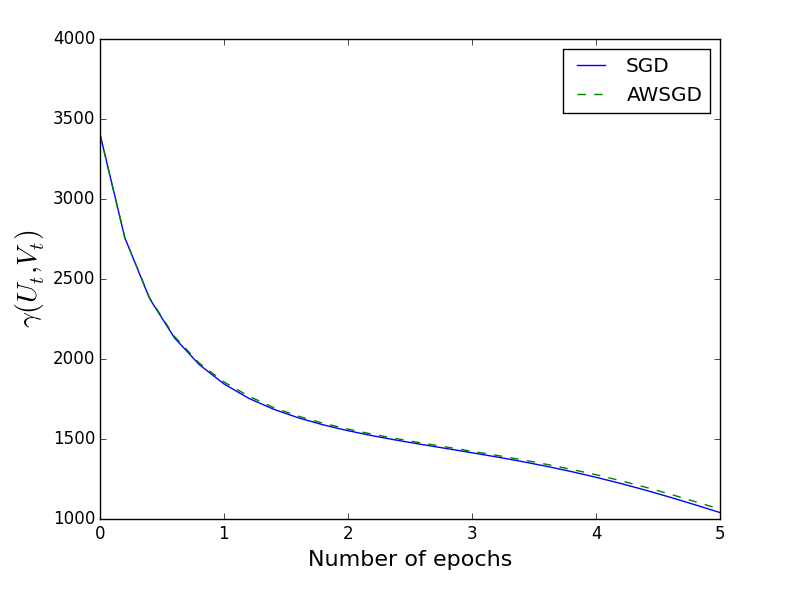}
    \caption{ Evolution of the training error as a function of the number of epochs
			on the simulated matrix with different access costs, with $f=5000$,
			for the uniformly-sampled SGD and AW-SGD using best $\rho_0$ for each algorithm.
	}
	\label{fig:numa_loss}
\end{figure}

\bibliographystyle{unsrt}
\bibliography{sgd}
\vspace{2cm}

%%%%%%%%%%%%%%%%%%%%%%%%%%%%%%%%%%%%%%%%%%%%%%%%%%%%%%%%%%%%%%%%%
\fi

\end{document}